\documentclass[10pt,twocolumn,letterpaper]{article}

\usepackage{cvpr}              % To produce the CAMERA-READY version
\usepackage[dvipsnames]{xcolor}

\definecolor{cvprblue}{rgb}{0.21,0.49,0.74}
\usepackage[pagebackref,breaklinks,colorlinks,citecolor=cvprblue]{hyperref}

\def\paperID{*****} % *** Enter the Paper ID here
\def\confName{CVPR}
\def\confYear{2024}

\title{When StyleGAN Meets Stable Diffusion: \\
a ${\mathcal{W}_+}$ Adapter for Personalized Image Generation}

\author{Xiaoming Li\qquad  Xinyu Hou\qquad Chen Change Loy\\
S-Lab, Nanyang Technological University\\
{\tt\small csxmli@gmail.com\qquad  xinyu.hou@ntu.edu.sg\qquad 
 ccloy@ntu.edu.sg}}

\usepackage{microtype}
\usepackage{color}
\usepackage{colortbl}
\usepackage{multirow}
\usepackage{makecell}
\usepackage{hhline}
\usepackage{cite}
\usepackage{caption}
\usepackage{bbding}
\usepackage{cuted}
\usepackage{comment}
\usepackage{flushend}

\begin{document}
\thispagestyle{empty}
\twocolumn[{%
	\renewcommand\twocolumn[1][]{#1}%
	\vspace{-1em}
	\maketitle
	\vspace{-0.2em}
	\begin{center}
		\centering
		\vspace{-0.34in}
		\includegraphics[width=.90\textwidth]{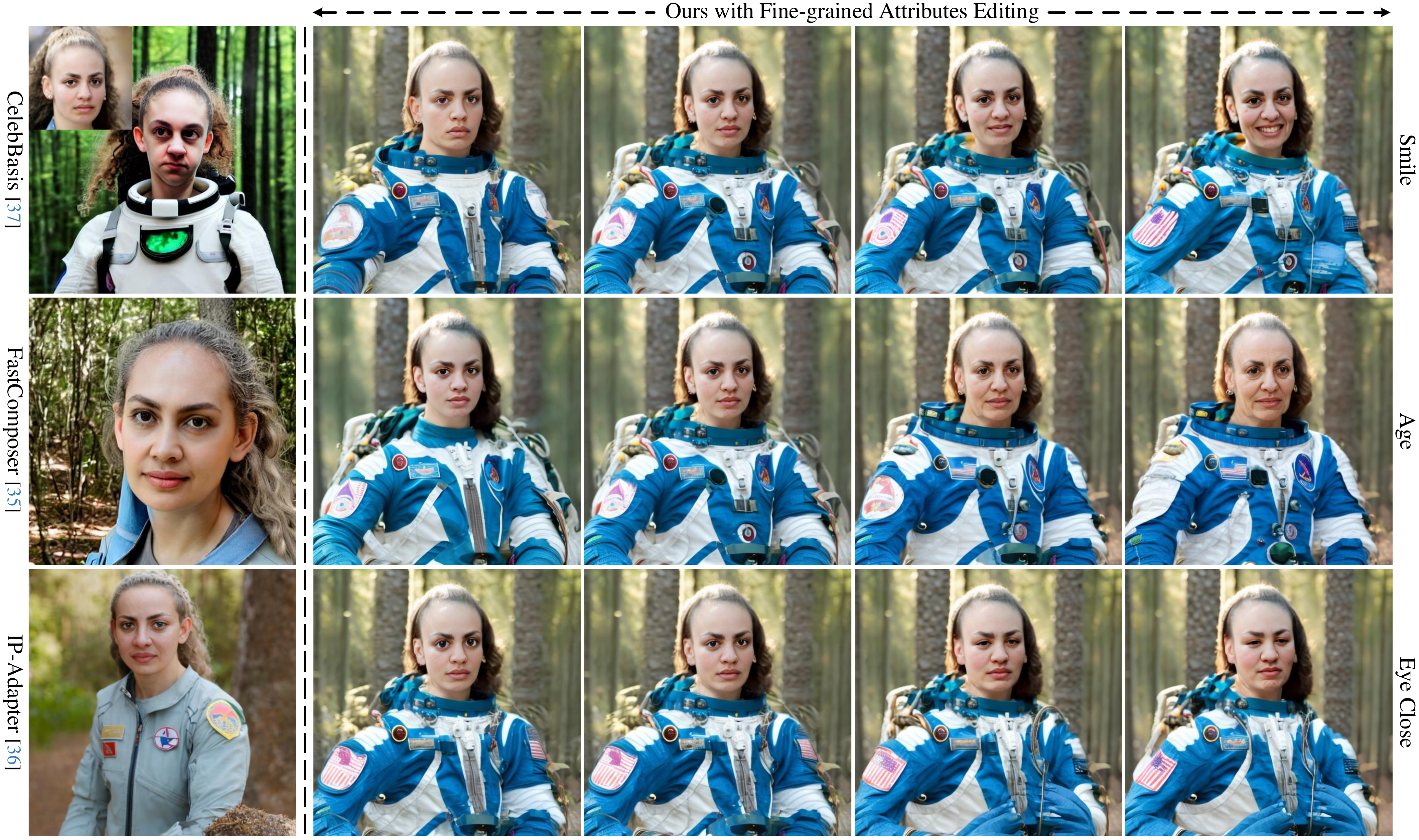}\vspace{0.1cm}
		\vskip -0.37cm
		\captionof{figure}{{Given a single reference image (thumbnail in the top left), our $\mathcal{W}_+$ adapter not only integrates the identity into the text-to-image generation accurately but also enables modifications of facial attributes along the $\Delta w$ trajectory derived from StyleGAN. The text prompt is ``a woman wearing a spacesuit in a forest''. 
}
		}
		\label{fig:fig1}
		\vspace{5pt}
	\end{center}%
}]

%--------------------------------------------------------
\begin{abstract}
\label{sec:abstract}
\vspace{-7pt}

Text-to-image diffusion models have remarkably excelled in producing diverse, high-quality, and photo-realistic images. This advancement has spurred a growing interest in incorporating specific identities into generated content. Most current methods employ an inversion approach to embed a target visual concept into the text embedding space using a single reference image. 
However, the newly synthesized faces either closely resemble the reference image in terms of facial attributes, such as expression, or exhibit a reduced capacity for identity preservation. 
Text descriptions intended to guide the facial attributes of the synthesized face may fall short, owing to the intricate entanglement of identity information with identity-irrelevant facial attributes derived from the reference image.
To address these issues, we present the novel use of the extended StyleGAN embedding space $\mathcal{W}_+$, to achieve enhanced identity preservation and disentanglement for diffusion models. By aligning this semantically meaningful human face latent space with text-to-image diffusion models, we succeed in maintaining high fidelity in identity preservation, coupled with the capacity for semantic editing. Additionally, we propose new training objectives to balance the influences of both prompt and identity conditions, ensuring that the identity-irrelevant background remains unaffected during facial attribute modifications.
Extensive experiments reveal that our method adeptly generates personalized text-to-image outputs that are not only compatible with prompt descriptions but also amenable to common StyleGAN editing directions in diverse settings. {Our source code will be available at \url{https://github.com/csxmli2016/w-plus-adapter}.}
\vspace{-3pt}
\end{abstract}
    
\vspace{-46pt}
\section{Introduction}
\label{sec:introduction}
\textit{What if we could be the protagonists of our own fantasies?} This paper primarily addresses personalized text-to-image (T2I) generation, a field that is attracting growing attention due to users' desire to craft their unique content. 
Overlapped with customized T2I~\cite{gal2022textual, ruiz2023dreambooth, wei2023elitea, kumari2023multiconcepta}, which integrates various visual concepts like human faces and objects into the generated imagery, our focus is on tailoring image generation to specific identities. This focus is motivated by applications such as story-boarding, where a consistent identity needs to be maintained across all images, despite variations in expressions or ages. Additionally, we believe that human faces possess unique, fine-grained intrinsic features that merit special attention and present a worthwhile area for exploration.

Recent personalized image generation approaches utilize a small reference set of a target identity and embed it into a specific space. The common choice of embedding space has been the textual embedding space used by large language models (LLMs). 
While existing methods based on textual embedding~\cite{gal2022textual, ruiz2023dreambooth, wei2023elitea} have proven capable of maintaining target identity, 
they are often limited by inherent trade-offs. Specifically, these methods face challenges in simultaneously preserving identity, generating varied facial attributes, and creating identity-irrelevant content that aligns with the text description.
We observe that these issues predominantly stem from the entangled nature of the textual embedding space, where a single pseudo word $\mathcal{S^*}$ struggles to distinctly isolate identity-related features from the reference image.
Efforts to separate such information include approaches like encoding a visual concept into multiple word embeddings~\cite{wei2023elitea} or employing a separate branch for extracting identity-irrelevant details~\cite{chen2023disenbootha}.
However, these methods tend to overlook the nuanced facial features critical to an individual's identity, resulting in incomplete disentanglement and suboptimal identity preservation.

In this study, we aim to more effectively separate identity-relevant and -agnostic features for better identity preservation, while also ensuring editability. 
To achieve this, we propose inverting the visual concept of a target identity into StyleGAN's \cite{karras2019style} $\mathcal{W}_+$ latent space, as opposed to using the textual embedding space. 
In particular, we introduce a mapping network to integrate the $\mathcal{W}_+$ space with the diffusion model and a residual cross-attention module to add the $w_+$ vector as an additional identity condition. 
This mapping network, once trained with \textit{(image, $w_+$)} pairs, can generalize to unseen individuals during inference without the need for a separate identity-specific model for each person.
We further present novel regularized training to ensure that edits do not alter identity-irrelevant regions and that the overall generation remains aligned with the prompt conditions.
It is noteworthy that although CelebBasis~\cite{yuan2023insertinga} explores a similar concept of creating a face-specific latent space through PCA of selected celebrity name embeddings, we argue that relying on such a limited dataset of celebrity names is insufficient for a comprehensive face latent space due to potential underrepresentation. Our experimental results confirm that our approach exhibits a superior capacity for identity preservation compared to CelebBasis.

The contributions of our work are summarised as follows: 1) We introduce extended StyleGAN $\mathcal{W}_+$ space as a target inversion latent space to better encode the identity-preserving facial concepts for personalized text-to-image synthesis. This is the first study that considers the fusion between StyleGAN $\mathcal{W}_+$ space and diffusion-based image generation. 2) Embedding target identity in $\mathcal{W}_+$ space enables smooth and controllable semantic editing on facial attributes in our text-to-image model. 3) %\cavan{
The effective disentanglement of identity-relevant and -irrelevant information in our model facilitates an identity-preserving generation that is not only diverse but also adaptable to a wide range of prompt instructions.

\section{Related Work}
\label{sec:related_work}

\textbf{StyleGAN Latent Space.} StyleGAN \cite{karras2019style} proposes to map input latent code to an intermediate latent space $\mathcal{W}$ to prevent warping of the training data distribution to fit in a particular probability distribution. The resulting $\mathcal{W}$ space is a semantically disentangled space that allows fine-grained controls over image synthesis. InterFaceGAN \cite{shen2020interpreting} and GANSpace \cite{harkonen2020ganspace} further interpret the latent space and identify several control directions such as age, gender, face angle, smile, etc. The powerful semantic editing ability has motivated research on inverting and editing real images in the StyleGAN space \cite{abdal2019image2stylegan, tov2021designing, pehlivan2022styleres}. Abdal \etal \cite{abdal2019image2stylegan} utilize a direct optimization framework to embed a given real image to the extended StyleGAN space. Tov \etal \cite{tov2021designing} deploy an encoder to perform the inversion and keep the concatenated latent codes close to the original StyleGAN space to maintain high perceptual quality and editability. Motivated by the disentanglement and editability of StyleGAN latent space, we aim to introduce it to the T2I diffusion models to achieve more controllable synthesis. We follow Image2StyleGAN to denote the extended $\mathcal{W}$ space represented by multiple nonidentical $w$ vectors as $\mathcal{W}_+$.

\noindent
\textbf{Personalized Image Synthesis.} Customization has been extensively studied in the context of T2I to generate images of specified objects or individuals. 
Gal \etal \cite{gal2022textual} directly apply learnable text embedding optimization. Ruiz \etal \cite{ruiz2023dreambooth} fine-tune the diffusion model for the target concept while preserving its class prior of {the concept}. Further improvements have been made to invert multiple concepts simultaneously with cross-attention fine-tuning \cite{kumari2023multiconcepta}, and disentangle background irrelevant information \cite{chen2023disenbootha}. Encoder-based approaches \cite{gal2023encoderbaseda, wu2023singleinsert, jia2023taminga, xiao2023fastcomposer, wei2023elitea, shi2023instantbootha, chen2023photoversea} are employed for their efficiency. The target visual concept is encoded to embedding space as additional conditions. SingleInsert \cite{wu2023singleinsert} adopts a two-stage scheme to insert concepts from a single image into the foreground region exclusively and enables the editing of the concept by text prompts. However, it is worth noticing that our method achieves more fine-grained editing with a single reference image at inference and does not necessitate identity-specific fine-tuned models.

With special attention on identity inversion utilizing the distinct features of human faces,
Yuan \etal \cite{yuan2023insertinga} define a celebrity space by applying PCA on selected celebrity name embeddings from CLIP and embedding new identities into the space via learnable coefficients. Valevski \etal \cite{valevski2023face0a} utilize pre-trained face recognizer as encoder and further project the face embedding to CLIP space. 
FaceChain \cite{liu2023facechain} trains separate face and style LoRAs \cite{hu2022lora} to synthesize specific faces in specific styles. They follow a data preprocessing pipeline and apply face fusion of the best reference face on the synthesized results. However, we observe that previous personalization approaches inherit a common limitation that the synthesized faces share similar facial attributes as the reference, indicating that the model cannot disentangle irrelevant information well and overfits the particular reference image(s).

\noindent
\textbf{Diffusion Model Adapters.} Light-weight adapters have been adopted to avoid the laborious work of fine-tuning large models and add additional controls \cite{zhang2023addinga, mou2023t2iadapter, ye2023ipadaptera} to the diffusion model. ControlNet \cite{zhang2023addinga} proposes to train task-specific adapters while freezing the original diffusion model to adapt to various input conditions. Concurrent work T2I-Adapter \cite{mou2023t2iadapter} constructs a lighter-weight adapter for controls on structure and color. Ye \etal \cite{ye2023ipadaptera} employ a decoupled cross-attention module to consider both text and image prompts in the denoising process. In our work, we consider identity information as an additional condition and align it with the T2I model for identity-preserving synthesis.
\section{Methodology}
\label{sec:methodology}

\begin{figure*}
    \setlength{\abovecaptionskip}{5pt} 
    \setlength{\belowcaptionskip}{-12pt}
    \centering
    \vspace{-15pt}
    \includegraphics[width=.98\textwidth]{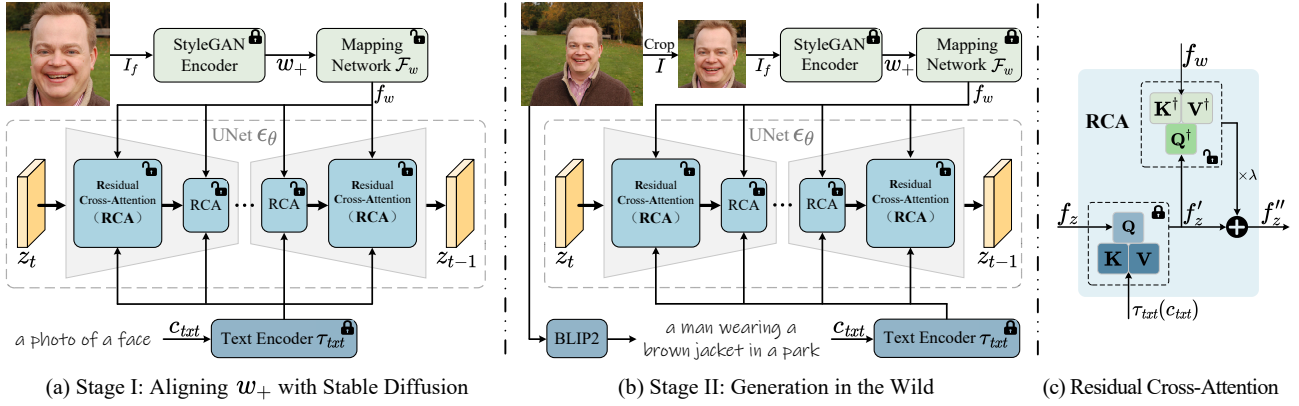}
    \caption{{Overview of $\mathcal{W}_+$ adapter training stages. \textit{Left}: Stage \uppercase\expandafter{\romannumeral1} for aligning $\mathcal{W}_+$ space with Stable Diffusion.  \textit{Middle}: Stage \uppercase\expandafter{\romannumeral2} for generating in-the-wild images with $w_+$ embeddings. \textit{Right}: details of residual cross-attention module. Open lock indicates trainable parts.}}
    \label{fig:stage1}
\end{figure*}

Our approach is capable of generating images that preserve identity while allowing for semantic edits, requiring just a single reference image for inference.
This capability is realized by innovatively aligning StyleGAN's $\mathcal{W}_+$ latent space with the diffusion model.
The training of our $\mathcal{W}_+$ adapter is divided into two stages. In Stage \uppercase\expandafter{\romannumeral1} (Sec.~\ref{sec:stage1}), we establish a mapping from $\mathcal{W}_+$ to SD latent space, using the resulting projection as an additional identity condition to synthesize center-aligned facial images of a specified identity.
In Stage \uppercase\expandafter{\romannumeral2} (Sec.~\ref{sec:stage2}), this personalized generation process is expanded to accommodate more dynamic, ``in-the-wild" settings, ensuring adaptability to a variety of textual prompts.

\subsection{Preliminary}
In our implementation, we use Stable Diffusion~\cite{rombach2022highresolution} as the foundational T2I model, which is one of the most commonly used latent space-based text-to-image generation model~\cite{van2017neural,esser2021taming}.
During the training phase, the model architecture includes: 1) a pre-trained encoder $\mathcal{E}(\cdot)$, which transforms an image $I$ into a latent representation with reduced dimensionality, 2) a conditional diffusion model tasked with predicting the latent code of a previous timestep, based on the text condition $c_\textit{txt}$ and the current timestep's latent code, and 3) a pre-trained decoder $\mathcal{D}(\cdot)$ that reconstructs the final latent code into the synthesized image. The learning objective of this model is formulated as follows:
%}
    \begin{equation}
        \small
    	\setlength{\abovedisplayskip}{5pt}
    	\setlength{\belowdisplayskip}{5pt}
    	\label{eqn:ldm}
        \mathcal{L}_\textit{LDM}=\mathbb{E}_{z_0, \epsilon\sim\mathcal{N}(0,1), t, c_\textit{txt}}\left[\left\|\epsilon-\epsilon_\theta\left(z_t, t, \tau_\textit{txt}(c_\textit{txt})\right)\right\|_2^2\right]\,
    \end{equation}
\noindent
where $z_0$ is $\mathcal{E}(I)$, $\epsilon$ is the ground truth noise added for current timestep $t$, $z_t$ is the latent code of timestep $t$. The $\tau_\textit{txt}$ denotes the pre-trained CLIP text encoder~\cite{radford2021learning} that converts the text prompt $c_\textit{txt}$ to textual embeddings. The $\epsilon_\theta$ represents the UNet \cite{unet} denosing network. In the inference stage, the denoising network is applied iteratively to denoise a random sampled noise $z_T$ to $z_0$ with condition $c_\textit{txt}$. The final result is then generated through the pre-trained decoder, \ie, $\mathcal{D}(z_0)$.

\subsection{Stage \uppercase\expandafter{\romannumeral1}: Aligning $\texorpdfstring{\mathcal{W}_+}{W+}$ with Stable Diffusion}
\label{sec:stage1}

The goal of Stage \uppercase\expandafter{\romannumeral1} is to align $\mathcal{W}_+$ space with SD to carry identity information in T2I generation while retaining its editing ability. To do so, we train a mapping network to project a $w_+$ embedding to SD latent space and inject this extra condition into SD by modifying the cross-attention module. The framework of Stage \uppercase\expandafter{\romannumeral1} is illustrated in Fig.~\ref{fig:stage1}. Details are introduced below.

\noindent 
\textbf{Training Pair Construction.} We adopt 100K discrete training samples to fit the continuous distribution of $\mathcal{W}_+$ and align it with SD. For each face image $I_f$, we use the pre-trained e4e~\cite{tov2021designing} encoder from StyleGAN inversion task to get its corresponding $w_+ \in \mathbb{R}^{18\times512}$ vector. The pairs of $\{I_f, w_+\}$ constitute our training data in this stage. Specifically, to generalize on real-world face images and improve $w_+$ diversity, two types of training pairs are built, \ie, synthetic face images from StyleGAN2~\cite{karras2020analyzing} and real-world face images from FFHQ~\cite{karras2019style}. Note that in Stage \uppercase\expandafter{\romannumeral1}, we only consider the aligned face images in order to exclude any extraneous influences.

\noindent
\textbf{Mapping Network.} Since the original $\mathcal{W}_+$ space is designated for StyleGAN generation, we project the vector $w_+ \in \mathbb{R}^{18\times512}$ to four tokens of dimension $768$ ($\mathbb{R}^{4\times768}$) to align with the input dimension of SD condition. The mapping network consists of four groups of linear layers and is denoted as $\mathcal{F}_w (\cdot)$. The latent space after mapping inherits the editability and disentanglement properties of $\mathcal{W}_+$ space and is compatible with the SD generation process.

\noindent
\textbf{Residual Cross-Attention.} 
To incorporate the identity condition from the projected $w_+$ embedding into the pre-trained SD model, we introduce a residual cross-attention module. In the standard SD framework, the output of cross-attention is determined using query features $f_z$ from the hidden state and the text condition $\tau_\textit{txt}(c_\textit{txt})$. The process is defined as
    \begin{equation}
        \small
        \setlength{\abovedisplayskip}{5pt}
        \setlength{\belowdisplayskip}{5pt}
        f^{\prime}_z=\operatorname{Attention}(\mathbf{Q}, \mathbf{K}, \mathbf{V})=\operatorname{Softmax}\left(\frac{\mathbf{Q K}^{\top}}{\sqrt{d}}\right) \mathbf{V}\,
    \end{equation}
where $\mathbf{Q}\!=\!f_z\mathbf{W}_q$, $\mathbf{K}\!=\!\tau_\textit{txt}(c_\textit{txt})\mathbf{W}_k$, $\mathbf{V}\!=\!\tau_\textit{txt}(c_\textit{txt})\mathbf{W}_v$ are the query, key, value matrices of the attention module, respectively. $\mathbf{W}_q$, $\mathbf{W}_k$, and $\mathbf{W}_v$ denote the corresponding projection matrices. The cross-attention module is where the latent noise interacts with text condition embeddings inside the denoising process. Following previous methods \cite{ye2023ipadaptera, wei2023elitea}, we add an additional condition by a separate cross-attention module. Our residual cross-attention is defined as:
    \begin{equation}
        \small
        \setlength{\abovedisplayskip}{5pt}
        \setlength{\belowdisplayskip}{5pt}
        f^{\prime\prime}_z=f^{\prime}_z + \lambda \cdot \operatorname{Attention}(\mathbf{Q}^{\dagger}, \mathbf{K}^{\dagger}, \mathbf{V}^{\dagger})\,
        \label{eqn:w-attn}
    \end{equation}
where $\mathbf{Q}^{\dagger}\!=\!f^{\prime}_z\mathbf{W}^{\dagger}_q$, $\mathbf{K}^{\dagger}\!=\!f_\textit{w}\mathbf{W}^{\dagger}_k$, and $\mathbf{V}^{\dagger}\!=\!f_\textit{w}\mathbf{W}^{\dagger}_v$. The $\mathbf{W}^{\dagger}_q$, $\mathbf{W}^{\dagger}_k$, and $\mathbf{W}^{\dagger}_v$ are the projection matrices for query, key, and value, respectively. The $f_w=\mathcal{F}_w (w_+)$ is mapped from $w_+$ by the mapping network. We use $\lambda$ as a scale parameter to balance the influence of text and identity conditions on the generation. We incorporate the decoupled cross-attention module in a residual fashion instead of the parallel approach as in previous works~\cite{ye2023ipadaptera, wei2023elitea} to avoid performance degradation on the original text condition. When $\lambda$ is set to 0, our $w_+$ vector has no impact on the pre-trained SD model. We set $\lambda$ to 1 during training. The $\mathbf{Q}^{\dagger}$, $\mathbf{K}^{\dagger}$ and $\mathbf{V}^{\dagger}$ are initialized from their corresponding $\mathbf{Q}$, $\mathbf{K}$, and $\mathbf{V}$, respectively. The residual cross-attention module is performed on all the cross-attention layers of SD. 

\noindent
\textbf{Learning Objectives.} The trainable parts in this stage include the mapping network $\mathcal{F}_w$ and $\{\mathbf{Q^{\dagger}}, \mathbf{K^{\dagger}}, \mathbf{V^{\dagger}\}}$ matrices in all residual cross-attention modules. The optimization target is formulated as:
    \begin{equation}
        \small
        \setlength{\abovedisplayskip}{5pt}
        \setlength{\belowdisplayskip}{5pt}
        \mathcal{L}_{w_+}=\mathbb{E}_{z_0, \epsilon, t, c_\textit{txt}, w_+}\left[\left\|\epsilon-\epsilon_\theta\left(z_t, t, \tau_\textit{txt}(c_\textit{txt}), \mathcal{F}_w(w_+)\right)\right\|_2^2\right]\,
        \label{eqn:w+}
    \end{equation}%
Eqn.~\eqref{eqn:w+} is similar to Eqn.~\eqref{eqn:ldm} except for the additional $\mathcal{F}_w(w_+)$ serving as identity condition and cross-attention structure modified inside $\epsilon_\theta$. The text prompt $c_\textit{txt}$ is randomly selected from several neutral templates describing a human face, \eg, ``a photo of a face'',  or ``a face'' (see suppl.).

Through mapping the $\mathcal{W}_+$ distribution into an embedding space compatible with SD, our approach effectively uses the $w_+$ vector to condition the personalized generation of aligned facial images. 
Furthermore, attribute directions $\Delta w$ derived from the original $\mathcal{W}_+$ space, encompassing features like smile, age, eye-opening, and others, remain applicable to our projected $w_+$ vector. 
This allows for personalized, fine-grained face editing, as demonstrated in Fig.~\ref{fig:fig1}.
%}

\subsection{Stage \uppercase\expandafter{\romannumeral2}: Generation in the Wild}
\label{sec:stage2}

In order to refine the performance of our $\mathcal{W}_+$-adapted SD model from Stage \uppercase\expandafter{\romannumeral1} for scenarios beyond controlled environments, we continue on a second stage of training. This phase is dedicated to further fine-tuning the weights of ${\mathbf{Q}^{\dagger}, \mathbf{K}^{\dagger}, \mathbf{V}^{\dagger}}$ across all residual cross-attention modules. To ensure that the projected $\mathcal{W}_+$ space exclusively encapsulates identity-related facial attributes, while remaining uninfluenced by irrelevant distractions 
from the background,
we keep the mapping network fixed during this stage. 
The architecture and workflow of Stage \uppercase\expandafter{\romannumeral2} are depicted in the middle of Fig.~\ref{fig:stage1}.

\noindent
\textbf{Training Data Construction.} For Stage \uppercase\expandafter{\romannumeral2}, in-the-wild images are used for training. For each image $I$, its corresponding text caption, $c_\textit{txt}$, is extracted using 
an off-the-shelf 
captioning tool~\cite{li2023blip}. The aligned face image $I_f$ is cropped from $I$, and $w_+$ is obtained from $I_f$ by the e4e encoder (same as Stage \uppercase\expandafter{\romannumeral1}). A face region mask, $M$, is also obtained from the image $I$, with 0 denoting the face region and 1 representing the non-face region.

\noindent
\textbf{Learning Objectives.} To keep high identity fidelity while encouraging high diversity of identity-irrelevant context, three losses are employed jointly in this stage. 

First, the reconstruction loss $\mathcal{L}_\textit{rec}$ is used to guide the denoising process:
    \begin{equation}
        \small
        \setlength{\abovedisplayskip}{5pt}
        \setlength{\belowdisplayskip}{5pt}
        \!\!\!\mathcal{L}_\textit{rec}\!=\!\mathbb{E}_{z_0, \epsilon, t, c_\textit{txt}, w_+}\left[\left\|\epsilon\!-\!\epsilon_\theta\left(z_t, t, \tau_\textit{txt}(c_\textit{txt}), \mathcal{F}_w(w_+)\right)\right\|_2^2\right]\,
        \label{eqn:rec}
    \end{equation}
\noindent
Note that Eqn.~\eqref{eqn:rec} is different from Eqn.~\eqref{eqn:w+} for that: 1) $z_0$ here is encoded from the in-the-wild image $I$ rather than the face image $I_f$, so the reconstruction goal differs, and 2) $c_\textit{txt}$ is the caption describing the in-the-wild image rather than a simple neutral prompt, therefore containing more information about the image context to facilitate reconstruction of $I$ with $w_+$ from $I_f$.

Second, when semantic edits $\Delta w$ are applied in the $\mathcal{W}_+$ space, the objective is to exclusively modify the facial attributes, leaving the remaining regions unchanged. In order to accomplish this, we propose a $w_+$ disentanglement loss to limit the editability of $w_+$ outside the face region:
    \begin{equation}
        \small
        \setlength{\abovedisplayskip}{5pt}
        \setlength{\belowdisplayskip}{5pt}
        \begin{aligned}
        \mathcal{L}_\textit{disen}=\|&M \cdot\epsilon_\theta(z_t, t, \tau_\textit{txt}(c_\textit{txt}), \mathcal{F}_w(w_+)) - \\
        & M\cdot \epsilon_\theta(z_t, t, \tau_\textit{txt}(c_\textit{txt}), \Psi(\mathcal{F}_w(w_+)))\|\,
        \end{aligned}
        \label{eqn:disen}
    \end{equation}
where $\Psi$ is the augmentation operation. During training, three augmentation strategies are adopted: 1) shuffle along the batch dimension of
$\mathcal{F}_w(w_+)$, 2) add random perturbations of Gaussian noise on $\mathcal{F}_w(w_+)$, and 3) combine both of them. By applying such constraints, augmented (or edited) $w_+$ vectors are forced to have similar non-face regions, thereby disentangling the effect of $\Delta w$ on the background.

Finally, since the identity condition $\mathcal{F}_w(w_+)$ should only influence the face region, which constitutes a relatively small proportion of the final output, we 
constrain its impact to be limited by a regularization loss:
    \begin{equation}
        \small
        \setlength{\abovedisplayskip}{5pt}
        \setlength{\belowdisplayskip}{5pt}
        \resizebox{.92\linewidth}{!}{$
        \!\!\!\mathcal{L}_\textit{reg}\!=\!\|M\cdot\epsilon_\theta\left(z_t, t, \tau_\textit{txt}(c_\textit{txt}), \mathcal{F}_w(w_+)\right) \!-\! M\cdot\epsilon_\theta\left(z_t, t, \tau_\textit{txt}(c_\textit{txt})\right)\|
        $}
        \label{eqn:reg}
    \end{equation}
\noindent
In this way, a more balanced effect of text and identity condition on the synthesized result is achieved, further mitigating the risk of unwanted noisy information introduced from $w_+$ and preserving the compatibility with text prompts.

The overall learning objective in Stage \uppercase\expandafter{\romannumeral2} is defined as:
    \begin{equation}
        \small
        \setlength{\abovedisplayskip}{5pt}
        \setlength{\belowdisplayskip}{5pt}
        \mathcal{L}= \mathcal{L}_\textit{rec} + \gamma_1 \cdot \mathcal{L}_\textit{disen} +\gamma_2 \cdot \mathcal{L}_\textit{reg}
        \label{eqn:stage2}
    \end{equation}
where $\gamma_1$ and $\gamma_2$ denote the trade-off parameters.
\vspace{-2pt}
\section{Experiments}
\label{sec:experiments}
%\vspace{-2pt}

\begin{figure*}[!t]
	\setlength{\abovecaptionskip}{5pt} 
	\setlength{\belowcaptionskip}{-10pt}
	\centering
	\vspace{-10pt}
	\includegraphics[width=.98\textwidth]{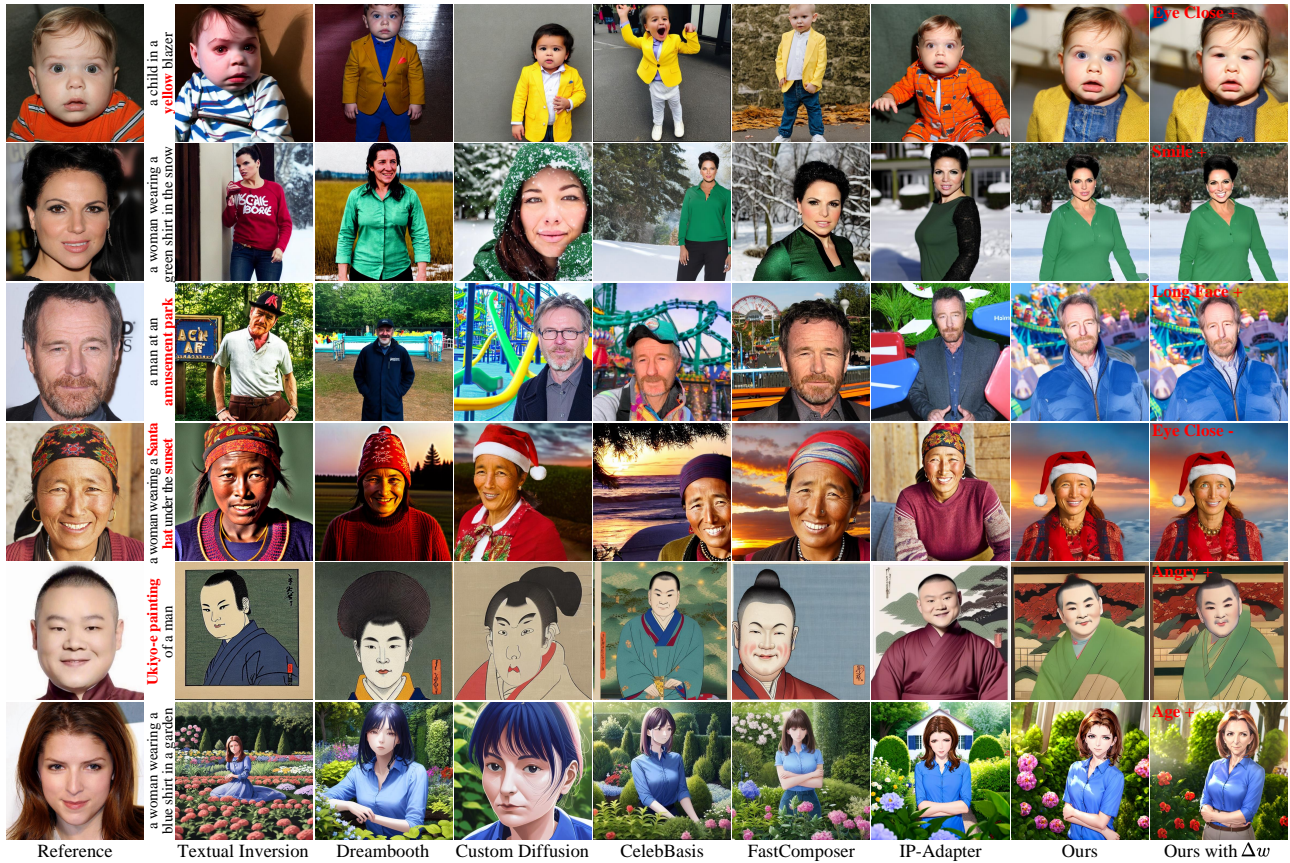}
	\caption{Visual comparisons with baselines on different scenarios of T2I generation. Best view by zooming in.}
	\label{fig:vis}
\end{figure*}

\noindent\textbf{Training Data.} In Stage \uppercase\expandafter{\romannumeral1}, we use the pre-trained e4e encoder~\cite{tov2021designing} to obtain the ${w}_+$ vector for each face image $I_f$ from FFHQ~\cite{karras2019style} and StyleGAN2~\cite{karras2020analyzing}. 
FFHQ dataset contains 70,000 images, among which 600 are split for validation while the rest are for training. We also synthesize 70,000 images using StyleGAN2.
In Stage \uppercase\expandafter{\romannumeral2}, we use FFHQ in-the-wild images (excluding those used for validation in Stage \uppercase\expandafter{\romannumeral1}) and SHHQ~\cite{fu2022stylegan} to optimize the $\mathcal{W}_+$ adapter. SHHQ is a human dataset that contains 40,000 high-quality full-body images. 
BLIP2~\cite{li2023blip} is introduced to generate the caption for each in-the-wild image. Face region mask $M$ is obtained based on the FFHQ alignment operation
and is eroded and blurred with kernel sizes of 32 and 7, respectively.

\noindent\textbf{Implementation Details.}
All training is conducted on a server with 8 Tesla V100 GPUs. The batch size is set to 16.
We employ the AdamW optimizer~\cite{loshchilov2017decoupled} with a learning rate of $1e{-4}$ and weight decay of 0.01. 
In Stage \uppercase\expandafter{\romannumeral2}, we adopt color jittering~\cite{zoph2020learning}, random rotation, and sampling for in-the-wild images to increase diversity. During training, the cropped images with incomplete or small faces are discarded. It takes around 24 hours to align the $\mathcal{W}_+$ space with SD in Stage \uppercase\expandafter{\romannumeral1}, and nearly 96 hours to train the $\mathcal{W}_+$ adapter in Stage \uppercase\expandafter{\romannumeral2}. 
In the inference, we adopt DDIM sampler~\cite{song2021denoising} with 50 steps. To enable classifier-free guidance~\cite{ho2022classifier}, we use the default settings and set the guidance scale to 7.5. We follow IP-Adapter~\cite{ye2023ipadaptera} to randomly drop the text $c_\textit{txt}$ and $w_+$ vector with a probability of 0.05.  $\gamma_1$ and $\gamma_2$ in Eqn.~\eqref{eqn:stage2} are set to 1.5 and 1.0, respectively.
The whole framework is implemented on Diffusers~\cite{diffusers}.
Our experiments are based on SD v1.5 and can be extended to other SD models (see suppl.).

\noindent

\begin{table}[t]
	\centering
	\vspace{6pt}
	\renewcommand\arraystretch{1.0}
	\small
	\setlength{\abovecaptionskip}{4pt}
	\setlength{\belowcaptionskip}{-5pt}
         \caption{Quantitative comparisons with previous methods. The best and second best results are highlighted by \textbf{bold} and \underline{underline}.
                 }
	\setlength{\tabcolsep}{2.3mm}
	{
		\begin{tabular}{l| c c c}
			\hline      
            Variants & CLIP Score$\uparrow$ & ID$\downarrow$  & Detection$\uparrow$ \\
            \hline
            Textual Inversion~\cite{gal2022textual}& .194  & .575 & .891\\
            Dreambooth~\cite{ruiz2023dreambooth}& .177  & .562 & .800\\
            Custom Diffusion~\cite{kumari2023multiconcepta}& .216  & .498 & .850\\
            FastComposer~\cite{xiao2023fastcomposer}& \underline{.265}  & {.419} & \underline{.950}\\
            IP-Adapter-Face~\cite{ye2023ipadaptera}& {.241} & \bf{.407} & \bf{.958}\\
            CelebBasis~\cite{yuan2023insertinga}& .253  & .448  & .916\\
            Ours & \bf{.267} & \underline{.418} & \underline{.950}\\
			\hline
	\end{tabular}}
	\label{tab:quan}
	\vspace{-12pt}
\end{table}

\subsection{Quantitative Comparison}
\noindent\textbf{Baselines.} The methods we evaluate can be categorized into two distinct groups: general object customization and specific face personalization. For general object customization, we use Dreambooth~\cite{ruiz2023dreambooth}, Textual Inversion~\cite{gal2022textual}, and Custom Diffusion~\cite{kumari2023multiconcepta}, implementing these using Diffusers~\cite{diffusers}. For specific face personalization, we select closely related works such as FastComposer~\cite{xiao2023fastcomposer}, IP-Adapter-Face~\cite{ye2023ipadaptera}, and CelebBasis~\cite{yuan2023insertinga}.
In our evaluation, we adhere to the settings established by CelebBasis~\cite{yuan2023insertinga} and FastComposer~\cite{xiao2023fastcomposer}. The metrics used for assessment include the {\bf{CLIP Score}}~\cite{radford2021learning}, which is calculated based on the average image-text similarity using CLIP-L/14, the Face {\bf{Detection}} Score determined using MTCNN~\cite{zhang2016joint}, and the Identity Distance ({\bf{ID}}) measured using ArcFace~\cite{deng2019arcface} on the detected facial regions.

Given the extensive fine-tuning requirements for each identity in methods like Dreambooth, Textual Inversion, Custom Diffusion, and CelebBasis – which can be notably time-consuming (for instance, Textual Inversion demands about an hour on a single V100 GPU server\footnote{{https://huggingface.co/docs/diffusers/training/text\_inversion}}) – we encounter constraints in evaluating a substantial number of test instances. 
To ensure a thorough and equitable quantitative analysis, we carefully choose 120 identities from the CelebA-HQ dataset~\cite{karras2017progressive}, using one reference image per identity for consistent comparisons. In addition, we randomly select 20 text prompts that describe various aspects such as clothing, styles, and backgrounds, to provide detailed characterizations of each individual.

Table~\ref{tab:quan} shows that optimization-based methods operating in the text embedding space, such as Textual Inversion, Dreambooth, and Custom Diffusion, fall short in accurately aligning with text prompts and preserving identity features. On the contrary, our method shows comparable performance to FastComposer and IP-Adapter-Face in terms of both CLIP Score and ID metrics. However, it is important to note that IP-Adapter has been fine-tuned on the large-scale LAION-2B dataset, whereas our model is trained on the relatively smaller FFHQ-wild and SHHQ datasets, comprising around 110,000 images.
Both FastComposer and IP-Adapter directly process the reference image as input, while our approach first maps it to the $\mathcal{W}_+$ space before embedding it into the SD model. This additional mapping step could potentially introduce minor discrepancies when integrating the reference image into the SD model, slightly affecting the ID results. Nevertheless, this slight trade-off in ID preservation is balanced by our method's ability to flexibly edit facial attributes while ensuring the background remains consistent. This unique balance of attribute editability and background consistency is a novel contribution that sets our method apart from previous approaches.

\subsection{Qualitative Comparison}

{Figure~\ref{fig:vis} depicts visual comparisons with baselines across various scenarios. Since Textual Inversion, Dreambooth, and Custom Diffusion are not specifically designed for facial images, we select results with the best facial quality from their generated set of  40 images. For the remaining methods, we choose from a set of 10. 
It is observed that Textual Inversion, Dreambooth, and Custom Diffusion 
encounter difficulties in capturing and maintaining identity details with a single reference, leading to suboptimal performance in this task.
CelebBasis and FastComposer face challenges in striking a balance between text compatibility and identity preservation. Though IP-Adapter shows improved identity retention, it tends to ignore text conditions in certain instances (see the 1\textit{st}, 3$\sim$5\textit{th} rows).}

Leveraging our $\mathcal{W}_+$ adapter, our approach successfully generates images that are not only compatible with text descriptions but also more effectively retain the target identity. Additionally, our method allows for the editing of facial attributes along the $\Delta w$ direction, causing only minor alterations in the non-facial regions (illustrated in the last column). Furthermore, our approach can be seamlessly adapted to other pre-trained SD models without the need for additional fine-tuning, while retaining its editing capabilities. This versatility is exemplified in the last row of Fig.~\ref{fig:vis}, which showcases our method's effectiveness with the dreamlike-anime model\footnote{https://huggingface.co/dreamlike-art/dreamlike-anime-1.0}.

\begin{figure}[!t]
	\setlength{\abovecaptionskip}{3pt} 
	\setlength{\belowcaptionskip}{-3pt}
	\centering
	% \vspace{-10pt}
	\includegraphics[width=0.96\linewidth]{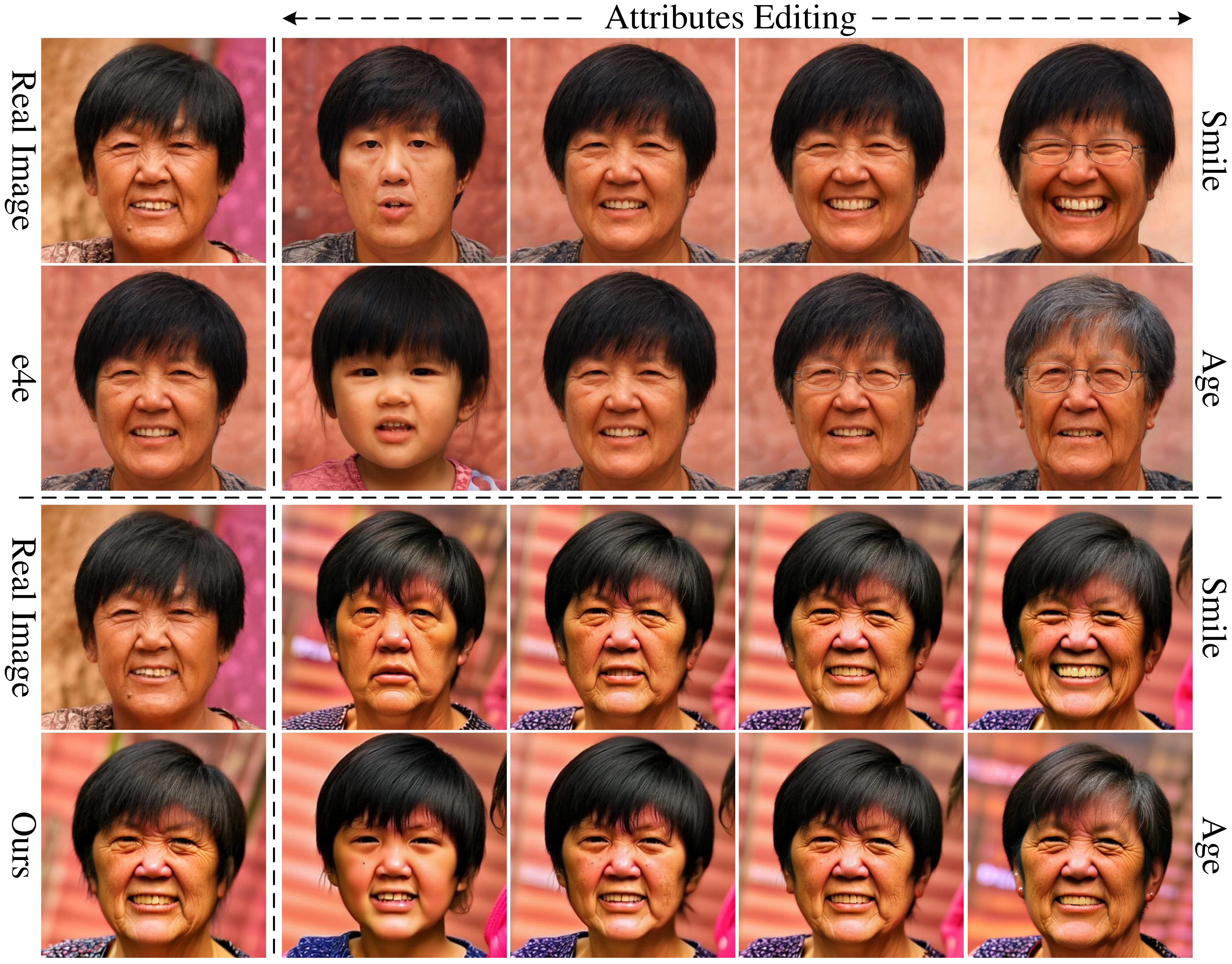}
	\caption{Inversion and editing comparisons between e4e and Ours in Stage \uppercase\expandafter{\romannumeral1}. They have the same $w_+$ and attributes editing $\Delta w$.}
	\label{fig:w}
\end{figure}

\subsection{Ablation Study}

\begin{table}[t]
	\centering
	\vspace{2pt}
	\renewcommand\arraystretch{1.0}
	\small
	\setlength{\abovecaptionskip}{2pt}
	\setlength{\belowcaptionskip}{-3pt}
         \caption{Quantitative comparisons of face inversion and editing. Editing (ID$\downarrow$) is the ID metric on results of $w_+$ and $w_+ - 3\cdot \Delta w$.}
	\setlength{\tabcolsep}{2.2mm}
	{
		\begin{tabular}{l| c c c | c c}
			\hline
            \multirow{2}{*}{\textbf{Methods}} & \multicolumn{3}{c|}{Inversion} & \multicolumn{2}{c}{Editing (ID$\downarrow$)} \\
            \hhline{~|-----}
            
            & ID$\downarrow$ & FID$\downarrow$  & LPIPS$\downarrow$ & Age & Smile \\
            \hline
            e4e~\cite{tov2021designing}& \textbf{.431} & 32.26 & \textbf{.205} & .409 &.312\\
            {Ours (Stage \uppercase\expandafter{\romannumeral1})} & .435 & \textbf{31.47} & .263 & \textbf{.393} & \textbf{.275}\\
			\hline
	\end{tabular}}
	\label{tab:w}
	\vspace{-12pt}
\end{table}

\textbf{Analyses of Aligning $\mathcal{W}_+$ in Stage \uppercase\expandafter{\romannumeral1}.} 
We examine if the StyleGAN's $w_+$ embedding is well aligned to the latent space of SD during Stage \uppercase\expandafter{\romannumeral1} training.
Fig.~\ref{fig:w} offers a qualitative comparison of inversion and attribute editing outcomes between e4e~\cite{tov2021designing} and our approach, applied to a real-world image. Both methods use the identical $w_+$ embedding generated by the e4e encoder.
As depicted in Fig.~\ref{fig:w}, our method yields inversion results on par with e4e, as seen in the left column, using the same $w_+$ vector. Furthermore, the alignment process we implement in this phase preserves the editability inherent to the $\mathcal{W}_+$ space. By introducing attribute editing directions $\Delta w$ (such as smile and age) from InterFaceGAN~\cite{shen2020interpreting}, our method can semantically adjust these attributes while maintaining the individual's identity. This is illustrated in the right columns, where the edited embedding, denoted as $w_++\alpha\cdot\Delta w$, reflects these changes. The parameter $\alpha$ is used to control the extent of attribute modification.

To quantitatively evaluate the alignment performance, we randomly select 1,000 images from CelebA-HQ~\cite{karras2017progressive} and measure their inversion results with metrics of ID, FID, and LPIPS~\cite{zhang2018unreasonable}\footnote{https://github.com/chaofengc/IQA-PyTorch}. 
As indicated in the Inversion column of Table~\ref{tab:w}, our method exhibits comparable performance to e4e in the image inversion task. 
For assessing attribute editing capabilities, we focus on two key attributes, namely smile, and age, applying a constant scale $\alpha$ (\ie, $w_+-3\cdot\Delta w$). The Editing column in Table~\ref{tab:w} demonstrates that our approach surpasses e4e in maintaining identity during the editing of facial attributes. These findings confirm that our method has successfully aligned the $\mathcal{W}_+$ space with the SD model, preserving not only inversion accuracy but also the robust editability of attributes.
\begin{table}[t]
	\centering
	\vspace{2pt}
	\renewcommand\arraystretch{1.0}
	\small
	\setlength{\abovecaptionskip}{2pt}
	\setlength{\belowcaptionskip}{-4pt}
	\caption{Quantitative comparisons of different variants.}
	\setlength{\tabcolsep}{3.2mm}
	{
		\begin{tabular}{l| c c c}
			\hline      
            Variants & CLIP Score$\uparrow$ & ID$\downarrow$  & Detection$\uparrow$ \\
            \hline
            Ours (\textit{PCA})& .243 & .526 & .936\\
            Ours (\textit{w/o} $\mathcal{L}_\textit{reg}$)& .136  & \bf{.461} & \bf{.951}\\
            Ours ($\mathcal{L}^{0.5}_\textit{reg}$)& .192  & .497 & .946\\
            {Ours (\textit{Full})} & \bf{.267}  & .516 & .943\\
			\hline
	\end{tabular}}
	\label{tab:var}
	\vspace{-12pt}
\end{table}

\begin{figure}[t]
	\setlength{\abovecaptionskip}{5pt} 
	\setlength{\belowcaptionskip}{-15pt}
	\centering
	\vspace{8pt}
	\includegraphics[width=0.98\linewidth]{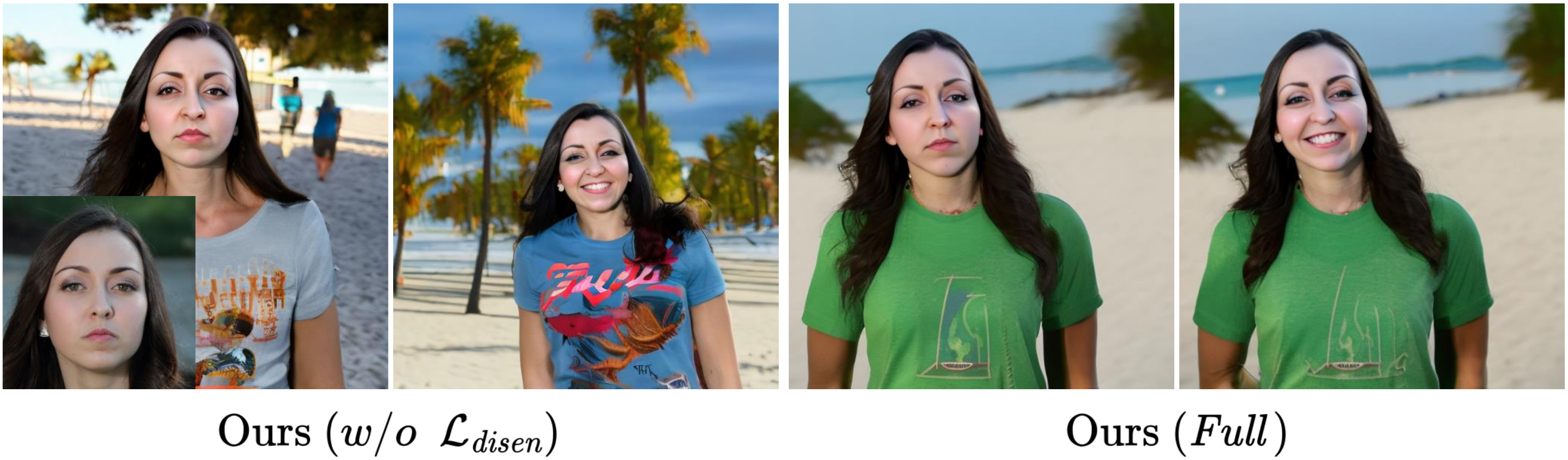}
	\caption{Visual comparisons of smile attribute editing. The prompt is ``a woman wearing a t-shirt on the beach''.}
	\label{fig:disen}
\end{figure}

\begin{figure*}[t!]
	\setlength{\abovecaptionskip}{5pt} 
	\setlength{\belowcaptionskip}{-12pt}
	\centering
	%\vspace{-10pt}
	\includegraphics[width=.968\textwidth]{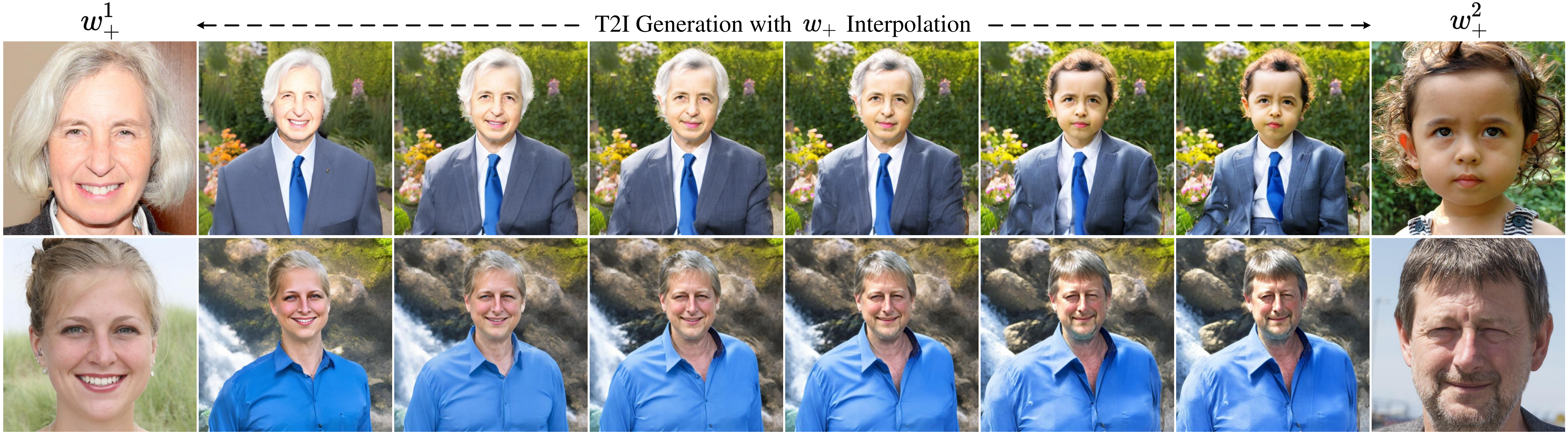}
	\caption{Visual results of $w_+$ embeddings interpolation during inference. The prompts are ``one person wearing suit and tie in a garden'' and ``one person wearing a blue shirt by a secluded waterfall'', respectively.}
	\label{fig:w_inter}
\end{figure*}

\begin{figure}[t]
	\setlength{\abovecaptionskip}{5pt} 
	\setlength{\belowcaptionskip}{-12pt}
	\centering
	%\vspace{-10pt}
	\includegraphics[width=0.938\linewidth]{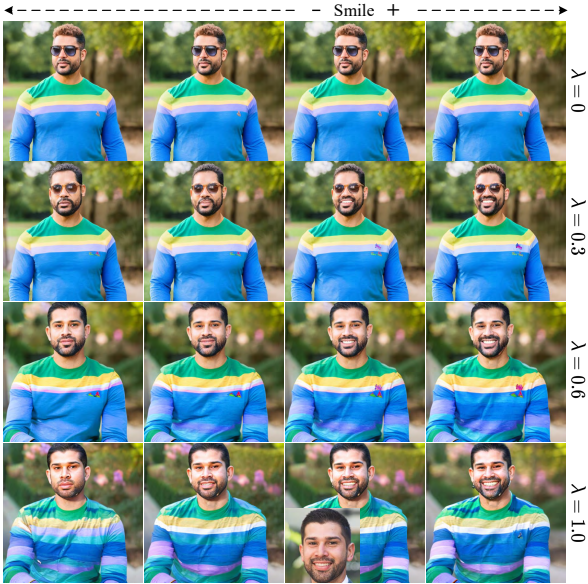}
	\caption{Visual results of using different $\lambda$ during inference. The prompt is ``a man wearing a rainbow shirt in a garden''.}
	\label{fig:lambda_inter}
\end{figure}

\noindent\textbf{Analyses of Variants.} 
We examine different variants to assess each component of our $\mathcal{W}_+$ adapter: 1) Ours~(\textit{PCA}), which substitutes our residual cross-attention with parallel cross-attention as in \cite{ye2023ipadaptera}, and 2) Ours (\textit{w/o} $\mathcal{L}_\textit{reg}$), which omits the $\mathcal{L}_\textit{reg}$ objective. The performance outcomes of these variants are presented in Table~\ref{tab:var}. It is observed that the residual cross-attention module outperforms the parallel cross-attention design.
We conjecture that in tasks focused on integrating local facial imagery into broader, in-the-wild scenarios, directly injecting facial embeddings into the original hidden state $f_z$ might adversely impact non-facial regions. Conversely, our residual cross-attention method determines fusion weights by calculating attention activations between the text-embedded hidden state $f^\prime_z$ and our facial embedding. This results in a more precise and effective fusion within the facial region.

Besides, the results suggest that the regularization loss $\mathcal{L}_\textit{reg}$ is crucial in maintaining semantic consistency between text prompts and images. In the absence of $\mathcal{L}_\textit{reg}$, the variant Ours~(\textit{w/o} $\mathcal{L}_\textit{reg}$) tends to overemphasize facial regions while neglecting the accompanying text descriptions, leading to a reduced CLIP score. Incorporating $\mathcal{L}_\textit{reg}$ with a weight of $\gamma_2=0.5$, as in Ours~($\mathcal{L}^{0.5}_\textit{reg})$, enhances the CLIP score but slightly compromises the ID metric. In comparison, Ours~(\textit{Full}) achieves a satisfactory trade-off in balancing the performance of ID and CLIP score. 

The impact of our disentanglement loss $\mathcal{L}_\textit{disen}$ (Eqn.~\eqref{eqn:disen}) is visually demonstrated in Fig.~\ref{fig:disen}. This example shows that omitting $\mathcal{L}_\textit{disen}$ leads to unintended alterations in non-facial regions when the smile editing direction $\Delta w$ is applied to $w_+$. This indicates the importance of $\mathcal{L}_\textit{disen}$ in effectively separating identity-relevant and irrelevant information. By combining our residual cross-attention, regularization loss, and disentanglement loss, Ours~(\textit{Full}) preserves identity and ensures compatibility with text prompts, even amidst changes to the face embedding.

\noindent\textbf{Analyses of $\lambda$ in Cross-attention.}
The parameter $\lambda$ determines the extent to which the $w_+$ vector influences the hidden state $f_z^\prime$. Fig.~\ref{fig:lambda_inter} shows the effects of varying $\lambda$ during inference. At $\lambda=0$, the outcome aligns with that of the original SD model, unaffected by any editing direction. However, as $\lambda$ increases, the generated identity more closely resembles the reference image. Interestingly, at relatively low values of $\lambda$ (for example, $\lambda=0.3$), even though the identity does not closely match the reference, our model effectively edits attributes in the specified direction. This observation suggests that our method proficiently leverages the $\mathcal{W}_+$ space from StyleGAN within the SD framework.

\noindent\textbf{Analyses of $w_+$ Interpolation.} 
We select two facial images and acquire their respective $w_+^1$ and $w_+^2$ embeddings from e4e encoder. The interpolated $w_+$ embedding is then obtained through $(1\!-\!\kappa)\cdot w_+^1 \!+\!\kappa\cdot w_+^2$, where $\kappa \in [0, 1]$. 
As shown in Fig.~\ref{fig:w_inter}, this T2I generation process results in a smooth transition in the facial regions of the generated images, while maintaining a similar and consistent layout throughout the interpolation.
This result confirms that the $w_+$ embedding in our method not only preserves the editability characteristic of the original StyleGAN but also effectively distinguishes between facial and background regions.

\subsection{Limitation}
Our work aims at integrating StyleGAN's editable $\mathcal{W}_+$ space into the SD model. We notice a challenge in this integration: the process of converting real images to $w_+$ vectors in StyleGAN often leads to a loss of detail, impacting the preservation of identity features. Despite employing a substantial number of training pairs $\{I_f, w_+\}$ to establish the mapping, we still observe limitations in maintaining identity fidelity and challenges in editing certain attributes (such as pose and glasses). The current framework of our method is designed to generate and edit images with only a single face. Our future work aims to explore the potential of applying localized injections of distinct $w_+$ embeddings to address multiple human subjects.

\section{Conclusion}
\label{sec:conclusion}
We have presented the first attempt to embed the $\mathcal{W}_+$ space of StyleGAN into the SD model. We showed that both the mapping network and the residual cross-attention module play crucial roles in facilitating the injection of $w_+$ embedding into the SD model, balancing between text prompt influence and identity conditions.
Our experiments demonstrate that the $\mathcal{W}_+$ space, as used in our approach, not only enables personalized text-to-image generation but also allows for precise editing of facial attributes. 
We envision this capability being highly beneficial in various practical applications, such as portrait customization with seamless attribute modifications. Furthermore, our methodology holds the potential for application across other object domains that possess distinct prior spaces.
%--------------------------------------------------------

{
    \small
    \bibliographystyle{ieeenat_fullname}
    \bibliography{main}
    
}

\clearpage
\setcounter{table}{0}
\setcounter{figure}{0}
\setcounter{section}{0}

\renewcommand\thesection{\Roman{section}}
\renewcommand\thefigure{\Alph{figure}}
\renewcommand\thetable{\Alph{table}}

\maketitlesupplementary

\noindent This supplemental material mainly contains:
\vspace{4pt}
\begin{itemize}
    %\vspace{-2pt}
    % \setlength{\itemsep}{2pt}
    %\setlength{\parsep}{-0pt}
    \setlength{\parskip}{2pt} 
    \item Details of mapping network $\mathcal{F}_w$ in Section~\ref{sec:mapping}
    \item Attribute editing in previous methods in Section~\ref{sec:pro}
    \item Analyses of two-stage training in Section~\ref{sec:stage}
    \item Visualization of residual cross-attention in Section~\ref{sec:vis}
    \item Text prompt for training Stage \uppercase\expandafter{\romannumeral1} in Section~\ref{sec:prompt}
    \item More generation and editing results in Figs.~\ref{fig:single} and \ref{fig:att}
    \item Performance on other SD models in Figs.~\ref{fig:otherSD1} and \ref{fig:otherSD2}
    \item More visual comparison with competing methods in Fig.~\ref{fig:compare}
\end{itemize}

\section{Details of Mapping Network}\label{sec:mapping}
\begin{figure}[h]
    \setlength{\abovecaptionskip}{9pt}
    \setlength{\belowcaptionskip}{-8pt}
    \centering
    \includegraphics[width=.45\textwidth]{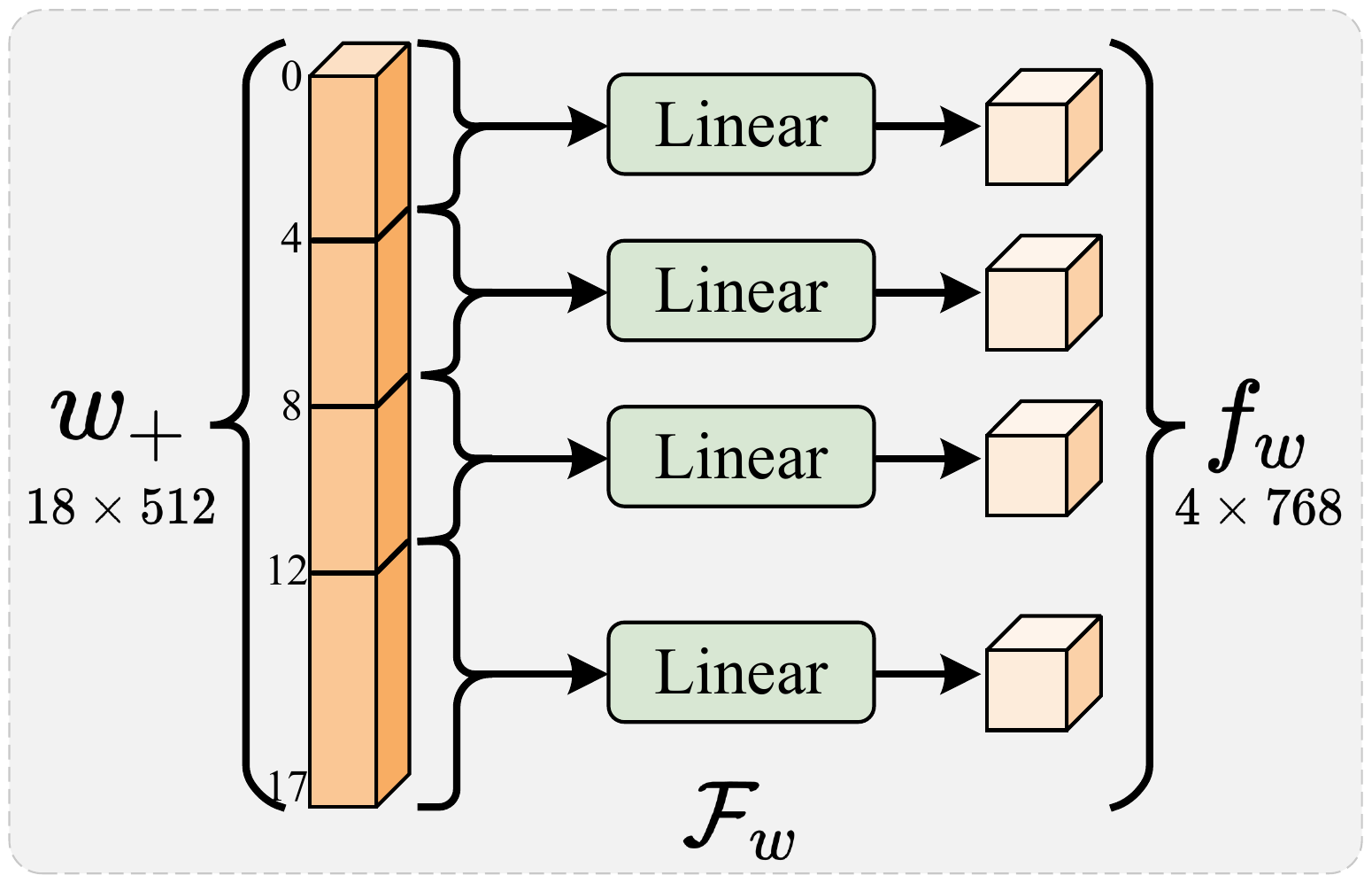} 
    \caption{Details of our mapping network $\mathcal{F}_w$.
    }
    \label{fig:mapping}
\end{figure}

The trainable modules of our $\mathcal{W}_+$ adapter consist of two parts, 1) the mapping network with 7.1 M parameters, and 2) residual cross-attention with 31.6 M parameters. Details of our mapping network $\mathcal{F}_w$ are shown in Fig.~\ref{fig:mapping}. To align with the input dimension of Stable Diffusion, the $w_+$ vector is divided into four groups. Each group is projected to a token of dimension 768 for Stable Diffusion V1.*.

\section{Attributes Editing in Previous Methods}
\label{sec:pro}

\begin{figure}[h]
    \setlength{\abovecaptionskip}{9pt}
    \setlength{\belowcaptionskip}{-13pt}
    \centering
    \includegraphics[width=.46\textwidth]{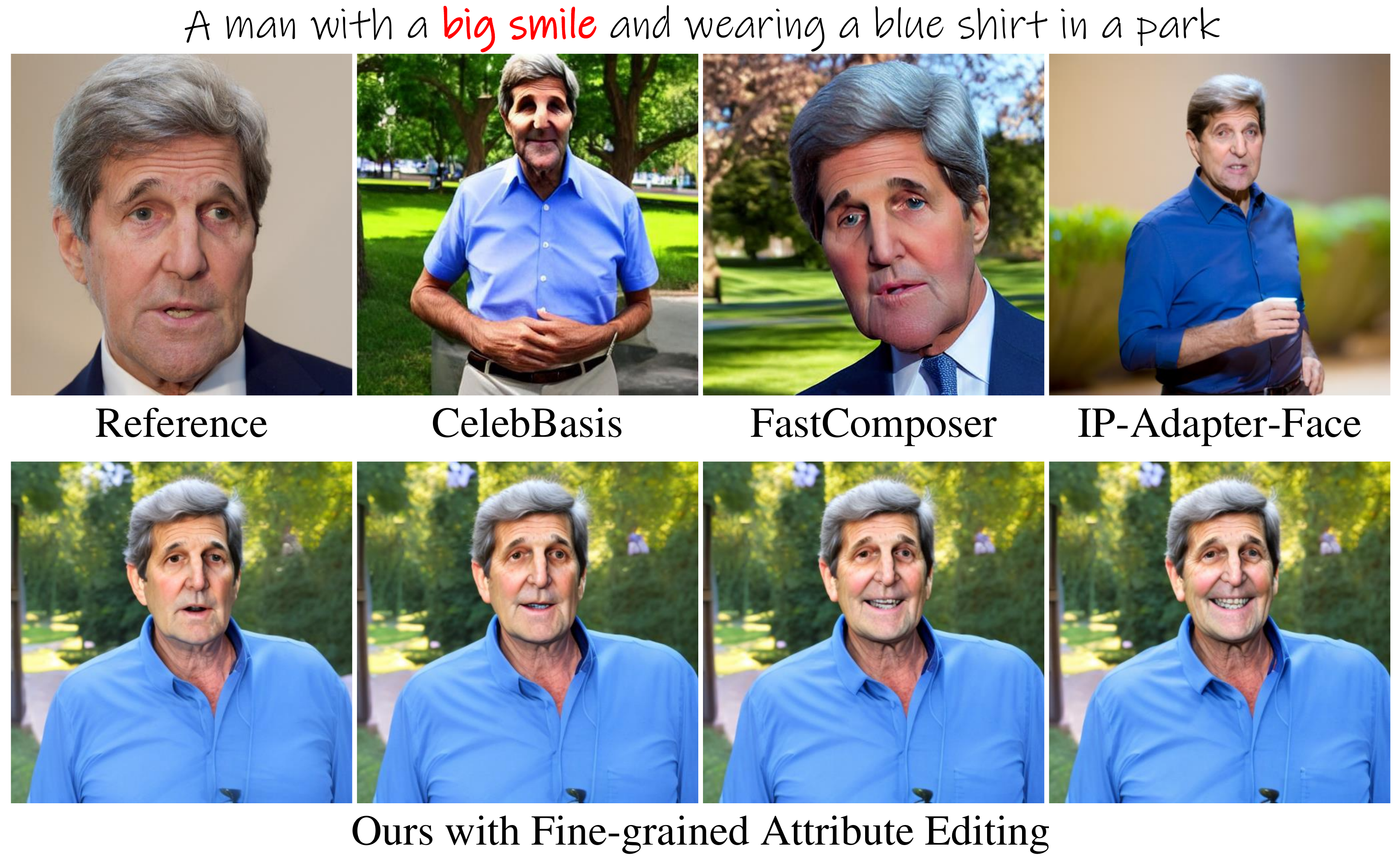} 
    \caption{Results of competing methods with text attributes.}
    \label{fig:pro}
\end{figure}

Fig.~\ref{fig:pro} shows that CelebBasis~\cite{yuan2023insertinga} encounters challenges in preserving the identity details. Both FastComposer~\cite{xiao2023fastcomposer} and IP-Adapter-Face~\cite{ye2023ipadaptera}) tend to overlook the facial attribute descriptions provided in the text captions. In contrast, our method not only excels in preserving identity but also edits common attributes with a seamless outcome.

\section{Analyses of Two-stage Training}
\label{sec:stage}
\begin{figure*}[h]
    \setlength{\abovecaptionskip}{6pt}
    \centering
    \includegraphics[width=.92\textwidth]{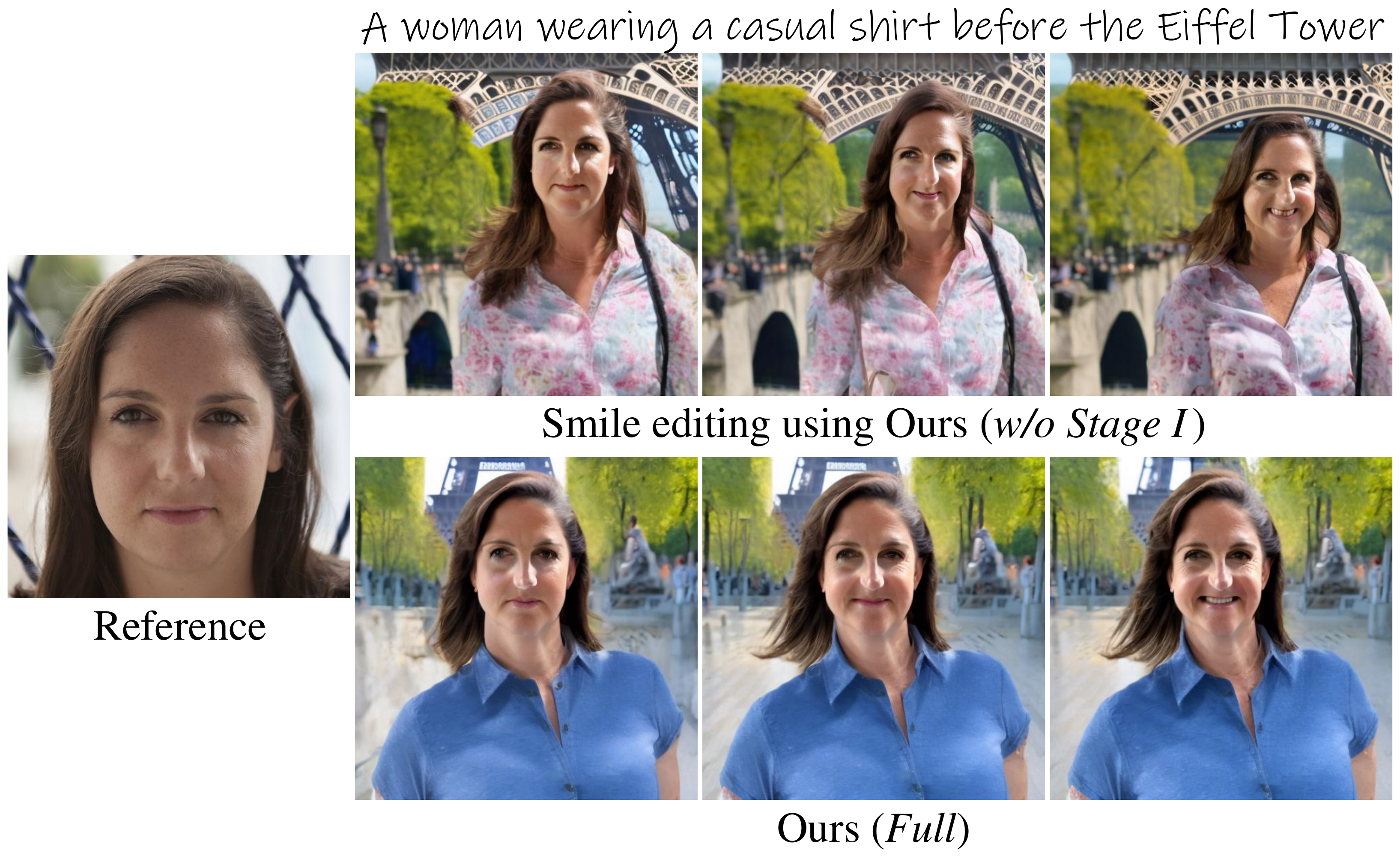} 
    \caption{Comparison between one- and two-stage training.
    }
    \label{fig:nostage1}
\end{figure*}

The training of our $\mathcal{W}_+$ adapter contains two stages. In Stage \uppercase\expandafter{\romannumeral1}, the model learns a mapping from $\mathcal{W}+$ to the SD latent space. Subsequently, in Stage \uppercase\expandafter{\romannumeral2}, the mapping network is fixed, and only the residual cross-attention is fine-tuned to facilitate in-the-wild generation. It is noteworthy that adopting a one-stage training approach, which directly optimizes the mapping network and residual cross-attention for in-the-wild generation, results in challenges in preserving consistent layout when editing attributes (see the 1-\textit{st} row in Fig.~\ref{fig:nostage1}).  This observation suggests that directly embedding the $w_+$ vector into diverse and complex in-the-wild generation may struggle to align well with the $\mathcal{W}_+$ space, thereby showing the necessity of our two-stage training approach.
%}

\section{Visualization of Residual Cross-attention}
\label{sec:vis}

\begin{figure*}[h]
    \setlength{\abovecaptionskip}{6pt}
    \setlength{\belowcaptionskip}{-2pt}
    \centering
    \includegraphics[width=.92\textwidth]{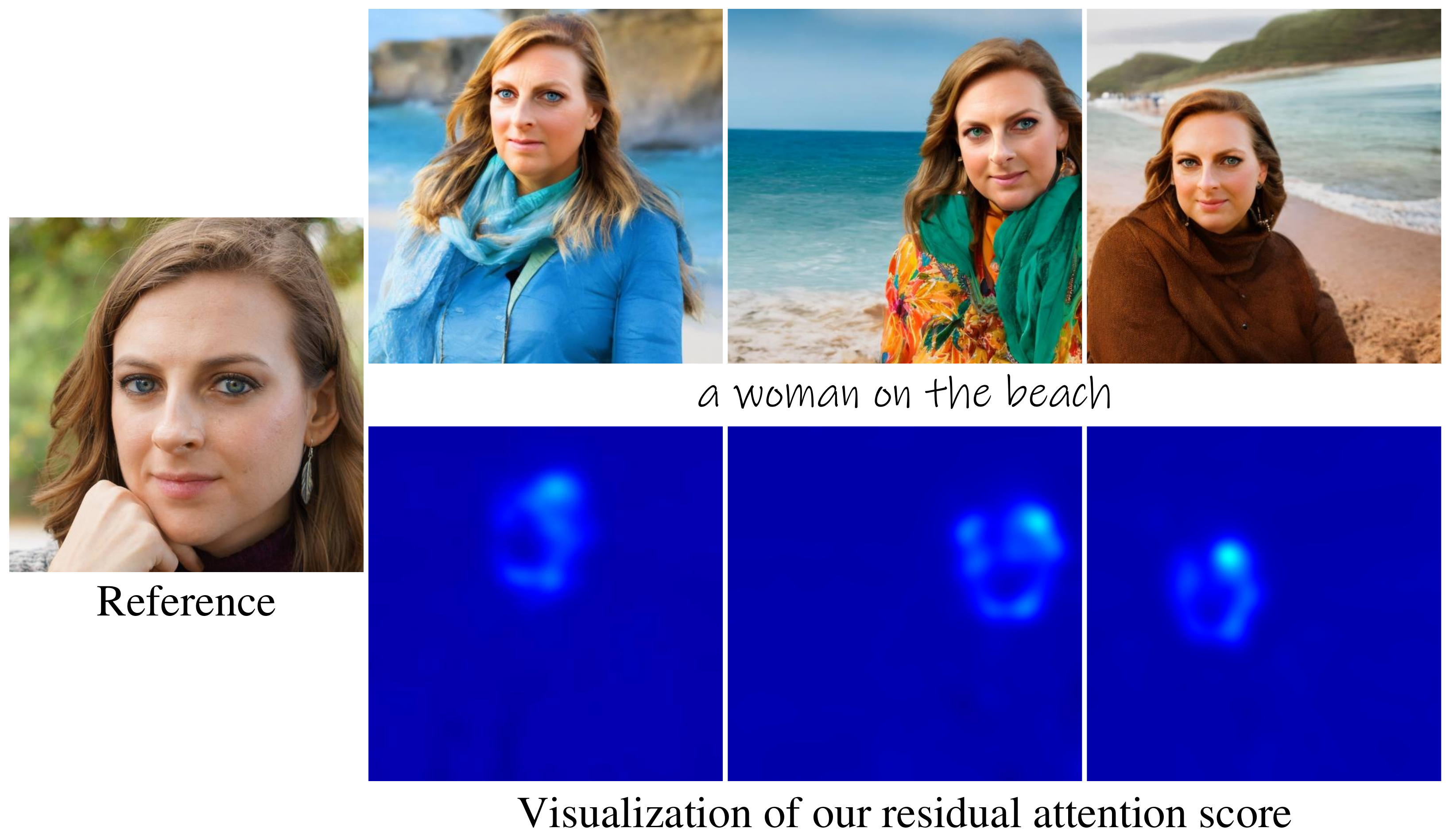} 
    \caption{Visualization of residual cross-attention score.
    }
    \label{fig:vis_residual}
\end{figure*}

\begin{figure*}[h]
    \setlength{\abovecaptionskip}{3pt}
    \centering
    \includegraphics[width=.93\textwidth]{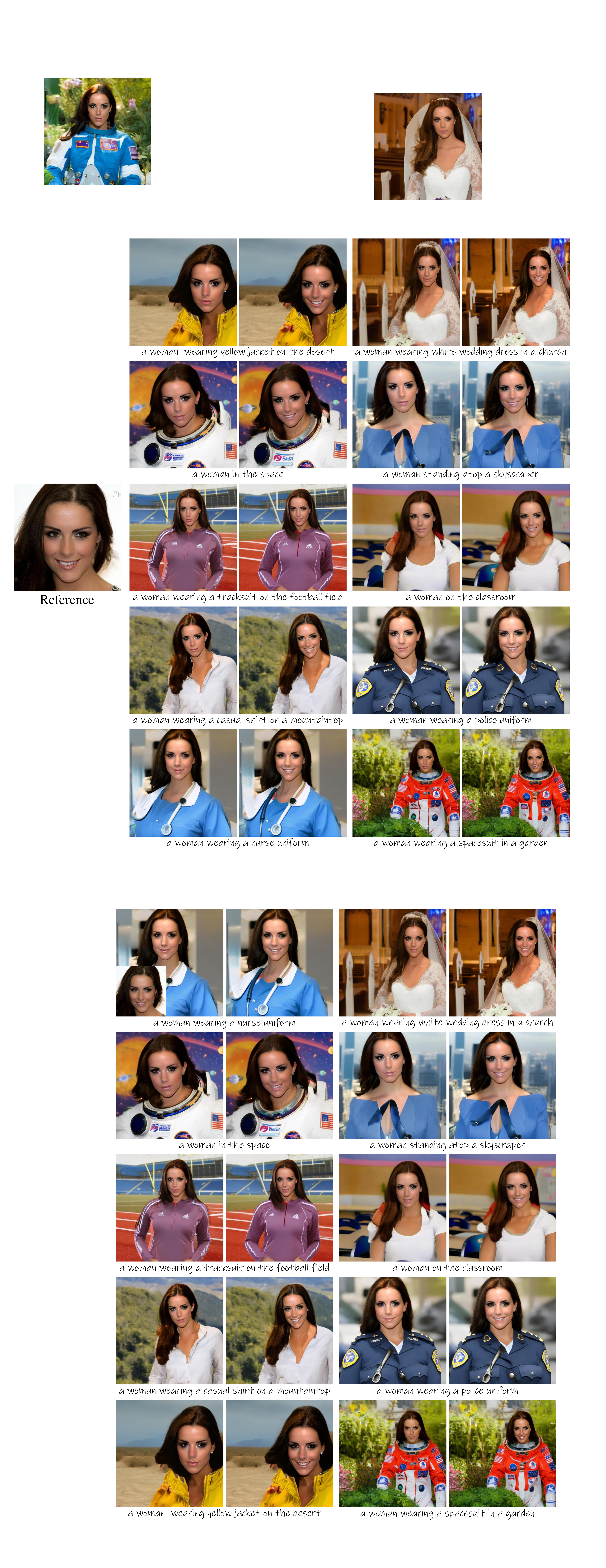} 
    \caption{More results of our $\mathcal{W}_+$ adapter for in-the-wild generation with different scenarios and attribute editing. 
    }
    \label{fig:single}
\end{figure*}

\begin{figure*}[h]
    \setlength{\abovecaptionskip}{3pt}
    \centering
    \includegraphics[width=.82\textwidth]{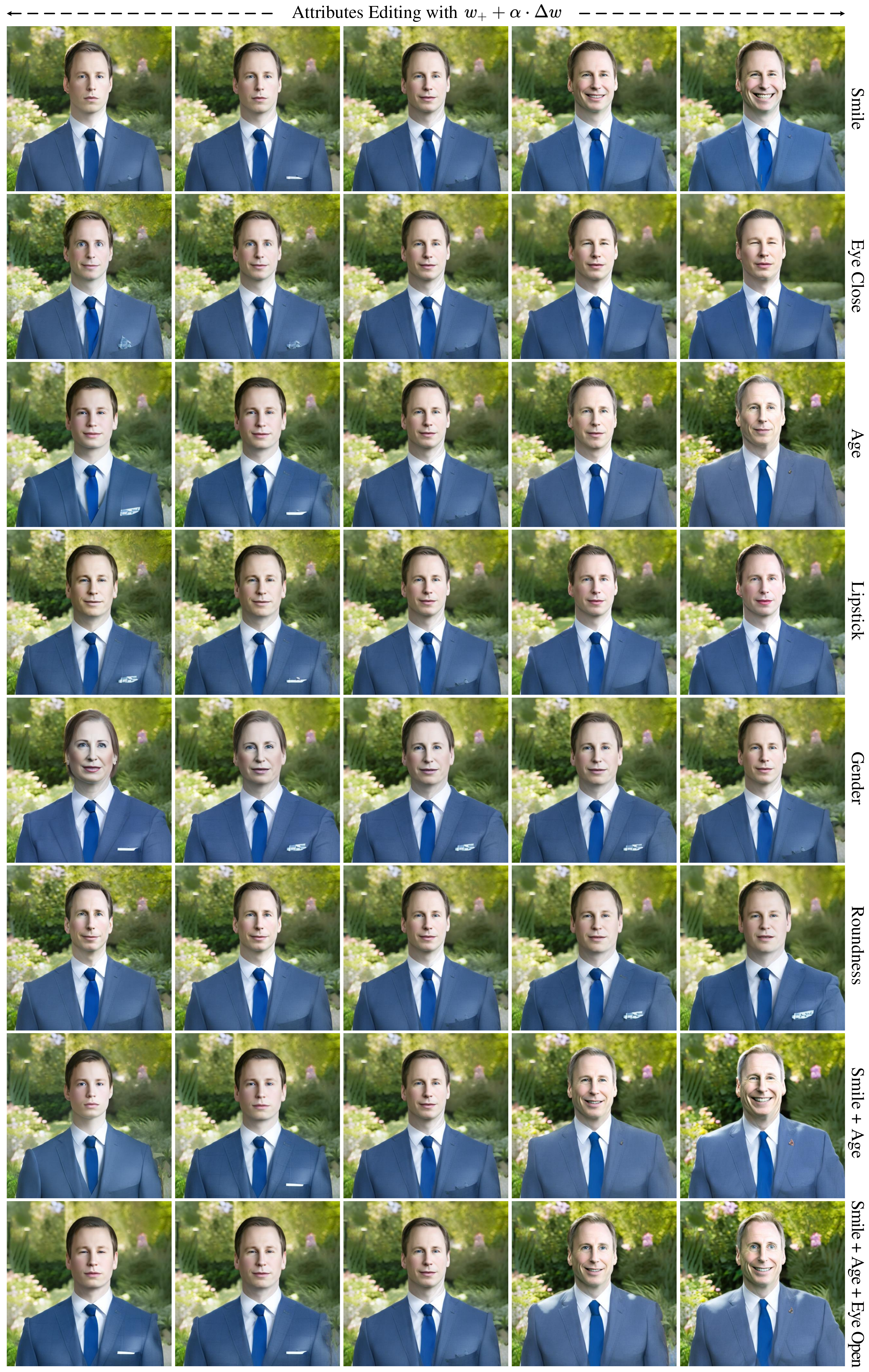} 
    \caption{More results of attribute editing for a single reference.
    }
    \label{fig:att}
\end{figure*}

\begin{figure*}[h]
    \setlength{\abovecaptionskip}{3pt}
    \centering
    \includegraphics[width=.77\textwidth]{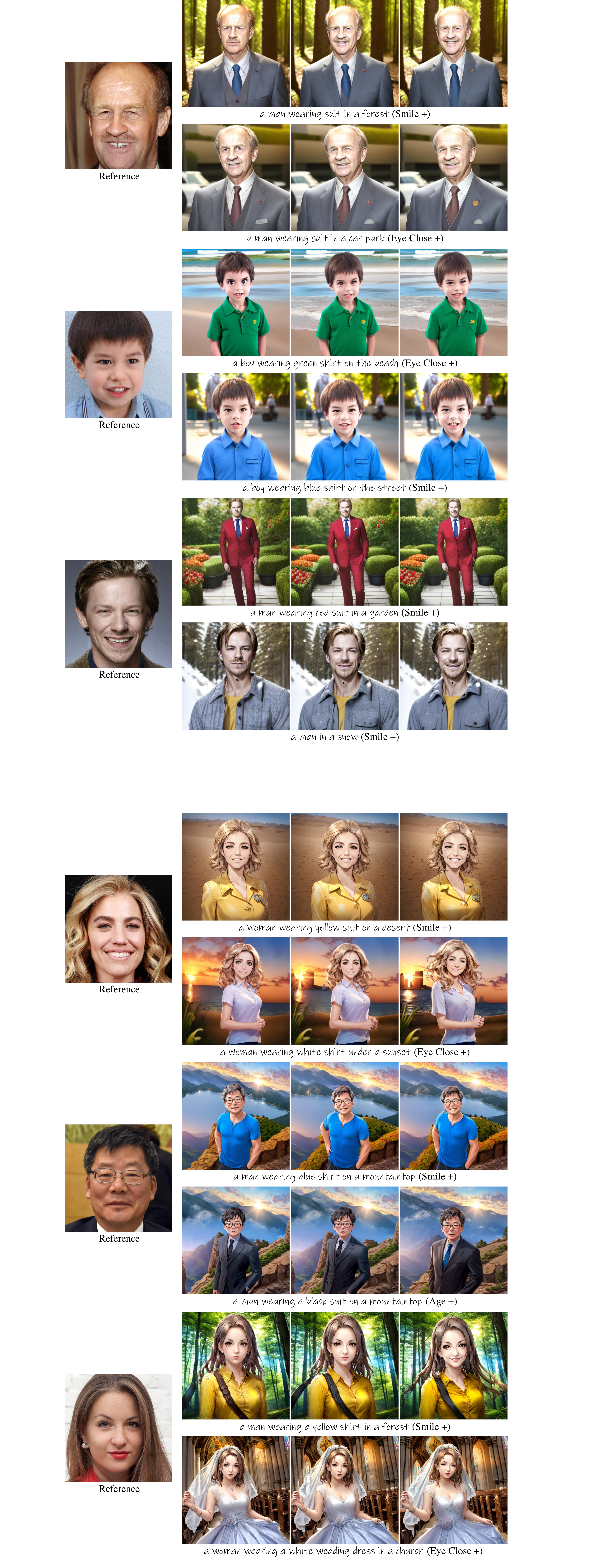} 
    \caption{Results of our $\mathcal{W}_+$ adapter using other SD model (\ie, Protogen).
    }
    \label{fig:otherSD1}
\end{figure*}

\begin{figure*}[h]
    \setlength{\abovecaptionskip}{3pt}
    \centering
    \includegraphics[width=.77\textwidth]{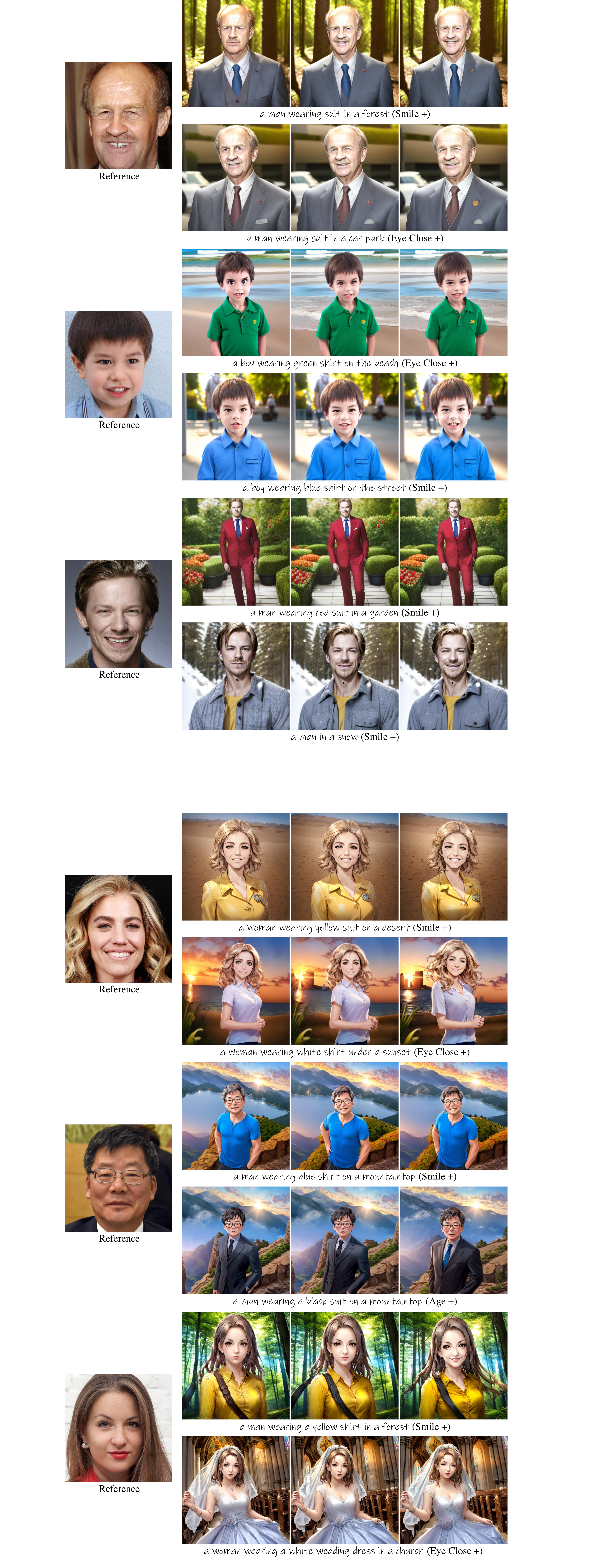} 
    \caption{Results of our $\mathcal{W}_+$ adapter using other SD model (\ie, dreamlike-anime).
    }
    \label{fig:otherSD2}
\end{figure*}

\begin{figure*}[h]
    \setlength{\abovecaptionskip}{3pt}
    \centering
    \includegraphics[width=.82\textwidth]{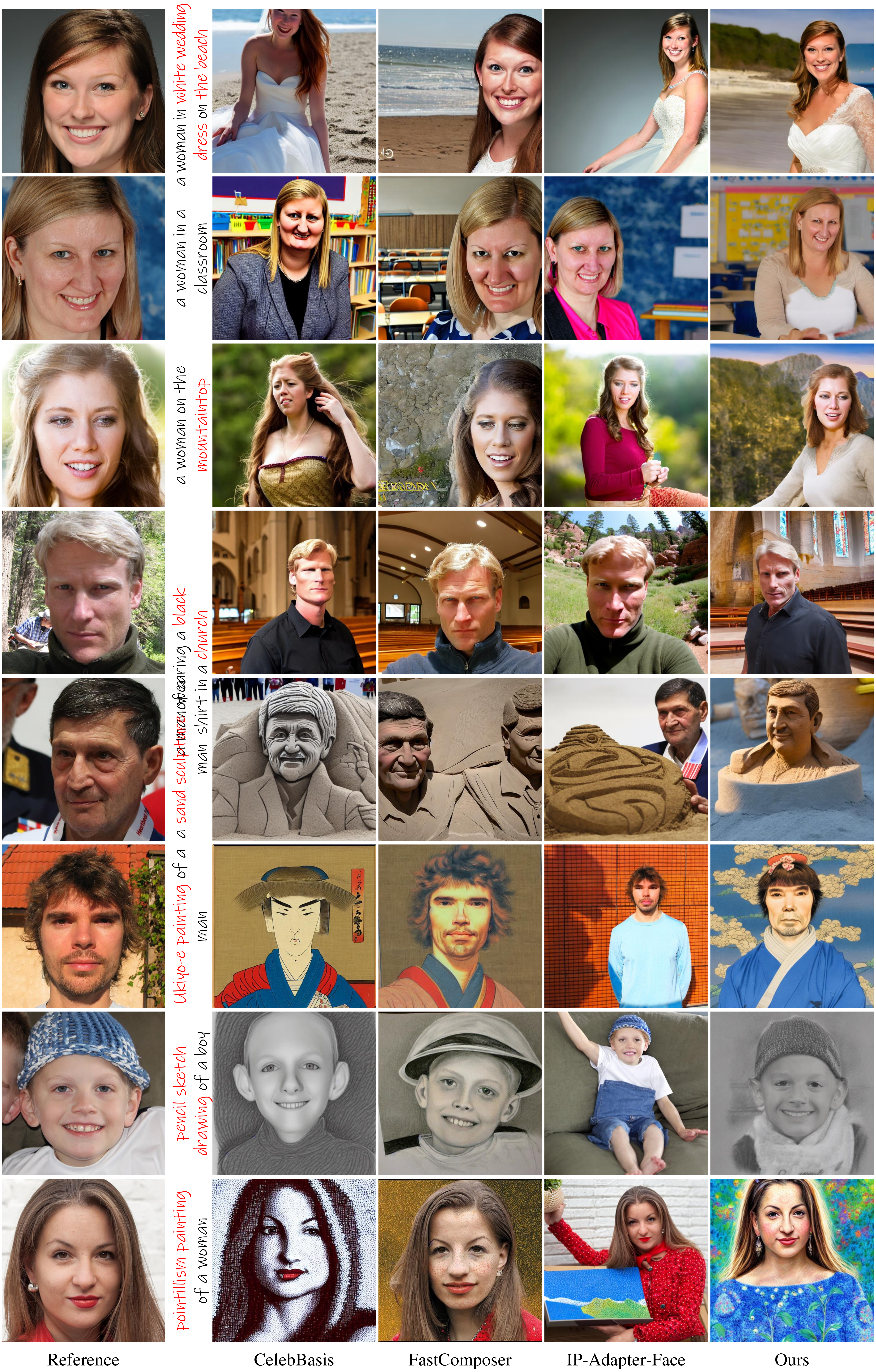} 
    \caption{More visual comparison with competing methods in different scenarios.
    }
    \label{fig:compare}
\end{figure*}

To demonstrate the influence of our residual cross-attention on the hidden states $f^{\prime}z$, we visualize the cross-attention score of the last layer by computing  $\operatorname{Softmax}\left(\frac{\mathbf{Q^\dagger K^\dagger}^{\top}}{\sqrt{d}}\right)$ at time step $t=1$. In Fig.~\ref{fig:vis_residual}, we observe an obvious impact of our $w+$ vector on the facial region, with minimal impact on the background. This observation showcases the effectiveness of our $\mathcal{W}_+$ adapter for editable face generation.

\section{Text Prompt for Training Stage \uppercase\expandafter{\romannumeral1}}
\label{sec:prompt}
The text prompt describing a human face in Stage \uppercase\expandafter{\romannumeral1} includes:
%\vspace{0pt}
\begin{itemize}
    \item a face
    \item a photo of a face
    \item a close-up of a face
    \item a depiction of a face
    \item a good photo of a face
    \item a photography of a face
    \item a cropped photo of a face
    \item a good photography of a face
    \item a close-up photography of a face
\end{itemize}

\end{document}